\documentclass[preprint,12pt]{elsarticle}

\usepackage{amssymb}
\usepackage{amsmath}
\usepackage{parskip}
\usepackage{multirow}
\usepackage{ocgx2}
\usepackage{media9}
\usepackage{pdfbase}
\usepackage{animate}
\usepackage{graphicx}
\usepackage{graphics}
\usepackage {framed}
\usepackage {amsxtra}
\usepackage {enumerate}
\usepackage{commath}
\usepackage{float}
\usepackage[caption = false]{subfig}

\usepackage{epstopdf}

\usepackage{pdfpages}
\usepackage{graphicx}
\usepackage{graphics}
\usepackage {framed}
\usepackage {amsxtra}
\usepackage {enumerate}
\usepackage{commath}
\usepackage[margin=1 in]{geometry}
\usepackage{float}
\usepackage[caption = false]{subfig}
\usepackage{bm}
\usepackage{amsthm}
\usepackage[utf8]{inputenc}
\usepackage[english]{babel}

\DeclareMathOperator*{\argminB}{argmin}   

\journal{arXiv}

\begin{document}

\begin{frontmatter}



\title{A novel nonconvex, smooth-at-origin penalty for statistical learning }


\author{Majnu John$^{a,}$$^{b,}$\footnote{Corresponding author, address: $350$ Community Drive, Manhasset, NY 11030, e-mail: {\sf mjohn5@northwell.edu, majnu.john@hofstra.edu}, Phone: +01\,718\,470\,8221, Fax: +01\,718\,343\,1659} Sujit Vettam$^{c,}$\footnote{The second author contributed to this work while he was a student at University of Chicago Booth School of Business. Currently employed in private sector.}, Yihren Wu$^{d}$}

\address{$^{a}$Departments of Mathematics and of Psychiatry, \\Hofstra University, Hempstead, NY.}
\address{$^{b}$Feinstein Institutes of Medical Research, \\Northwell Health System, Manhasset, NY.}
\address{$^{c}$The University of Chicago Booth School of Business,\\Chicago, IL.}
\address{$^{d}$Department of Mathematics, \\Hofstra University, Hempstead, NY.}

\begin{abstract}

 Nonconvex penalties are utilized for regularization in high-dimensional statistical learning algorithms primarily because they yield unbiased or nearly unbiased estimators for the parameters in the model. Nonconvex penalties existing in the literature such as SCAD, MCP, Laplace and arctan have a singularity at origin which makes them useful also for variable selection. However, in several high-dimensional frameworks such as deep learning, variable selection is less of a concern. In this paper, we present a nonconvex penalty which is smooth at origin. The paper includes asymptotic results for ordinary least squares estimators regularized with the new penalty function, showing asymptotic bias that vanishes exponentially fast. We also conducted an empirical study employing deep neural network architecture on three datasets and convolutional neural network on four datasets. The empirical study showed better performance for the new regularization approach in five out of the seven datasets.

\end{abstract}


\end{frontmatter}





\noindent

\section{Introduction}

Regularization is often employed to reduce overfitting in statistical learning algorithms. There are several approaches for regularization of high-dimensional statistical models. One commonly employed approach appends a penalty function to the loss function (i.e. the cost function) of the model to regularize the parameters in the model. Widely used penalty functions include Ridge (Hoerl and Kennard 1970; Frank and Friedman 1993), Lasso (Tibshirani 1996), Bridge\footnote{Lasso and Ridge are special cases of the class of Bridge penalty functions, $L_{q}, 0 < q < \infty$; $q = 1$ for Lasso and $q = 2$ for Ridge. However, because of the importance of these two special penalty functions in theory, methodology and applications, we will always emphasize them whenever we mention $L_{q}, q \geq 1$.} (Fu 1998), Elastic Net (Zou and Hastie 2005), SCAD (Fan and Li 2001), MCP (Zhang 2010), and penalty functions useful for specific contexts such as group-Lasso (Yuan and Lin 2006), fused-Lasso (Tibshirani, Saunders, Rosset and Knight 2005), adaptive-Lasso (Zou 2006) and several others. Bayesian versions of the above penalty functions (Park and Casella 2008) including a recent Bayesian penalty called Hierarchical Adaptive Lasso (Seto, Wells, Zhang 2021) specifically suited for deep neural networks have also appeared in the literature. Two other relatively new penalties that have appeared in the literature are Laplace and arctan penalties (Trzasko and Manduca 2009; Wang and Zhu 2016; Vettam and John 2022). Substantial development of the field in the last two decades includes studies related to optimization algorithms and theoretical aspects of penalty based regularization, and it is literally impossible to summarize this rich and broad field in a few introductory paragraphs.

Various penalty functions such as Ridge, Lasso, Bridge, Elastic Net, SCAD and MCP have distinct features that serve unique purposes. Broadly speaking, the penalty functions may be classified as convex (e.g. Elastic Net and Bridge, $L_{q}, q \geq 1$ - including Lasso and Ridge) versus nonconvex (e.g. SCAD, MCP, Laplace, arctan and $L_{q}, 0 < q < 1$). Another classification is based on whether the penalty functions are smooth at origin (e.g. Bridge, $L_{q}, q > 1$ - including Ridge) versus whether they have a singularity at origin (e.g. Elastic Net, SCAD, MCP and Bridge, $L_{q}, 0 < q \leq 1$ - including Lasso). Penalties with singularity at origin could also be used for variable selection (that is, they set many parameters to zero) while as penalties which are smooth at origin do not have this property. Convex penalties are obviously better for practical optimization implementation. However, convex penalties yield biased estimates of the parameters. On the other hand, parameter estimation based on nonconvex penalties satisfy the unbiasedness or the near unbiasedness property. Hence if bias is an issue, it is advantageous to employ nonconvex penalties\footnote{Note that bias is not always an issue because if prediction is of main concern then often biased (i.e. shrinkage) estimators could substantially reduce the variance thereby lowering the prediction error.}.

The nonconvex penalties seen so far in the literature, SCAD, MCP, Laplace, arctan and $L_{q}, 0 < q < 1$, are all singular at the origin. Singularity at origin, although necessary when the penalty function is used for variable selection, could pose theoretical and computational challenges when the focus is on regularization alone without variable selection. Nonconvex Bridge functions ($L_{q}, 0 < q < 1$) for example have unbounded derivatives at origin making zero always a local minimum for the composite (cost + penalty) objective function thereby preventing estimation error from getting smaller beyond a threshold. Even for nonconvex penalties such as SCAD, MCP, Laplace and arctan with bounded derivatives at origin, the singularity at origin could pose optimization challenges. In addition to such optimization challenges seen in practice, there are theoretical disadvantages too, especially when establishing the consistency of the stationary points for the objective functions. It has been noted in the literature (Qi, Cui, Liu and Pang 2021) that objective functions which are a sum of a nonconvex, singular-at-origin penalty function coupled with either a smooth nonconvex loss function or a nonsmooth convex loss function, often fails to satisfy a theoretical condition known as `Clarke regularity'. Theoretical justifications such as consistency of the stationary solutions are very challenging for non-Clarke regular problems especially since the uniform law of large numbers for random set-valued Clarke regular mappings established in Shapiro and Xu, 2007 cannot be used.

All these considerations motivated us to look for a penalty function which behaves like $L_{2}$ near the origin (that is, smooth and convex near origin) but behaves like nonconvex penalties away from the origin, for large values of the parameter. In this paper, we present such a penalty (see Eq. (2)). We label the new penalty as `Gaussian penalty' firstly as a homage to the pioneering German mathematician and secondly also because the form of the new penalty function in the positive quadrant has many similarities to the Gaussian distribution function.

The paper is organized as follows. In the next section we introduce required background notation and present the new penalty function. Section 3 focuses on asymptotic consistency results when the loss function is the least squares function considered for ordinary linear regression. In section 4 we assess the empirical performance of the new regularization approach on data examples employing deep neural networks and convolutional neural networks. We summarize and briefly discuss in the final section.

\section{Background, notation, and the new penalty function}

 For an input random vector $X$ and a real-valued or binary output variable $Y$ in a statistical learning framework, consider a function $f_{\pmb{\bm{\beta}}}(X)$, parameterized by $\pmb{\bm{\beta}}$, used for predicting $Y$ given the values of the input $X$. The optimal function $f_{\hat{\pmb{\bm{\beta}}}}(X)$ that minimizes the prediction inaccuracies is chosen by minimizing a loss function $\mathcal{L}(Y, f_{\pmb{\bm{\beta}}}(X))$ (Hastie, Tibshirani and Friedman, 2001). For example, employing the squared error loss, $\mathcal{L}(Y, f_{\pmb{\bm{\beta}}}(X)) = (Y - f_{\pmb{\bm{\beta}}}(X))^{2}$ leads to minimizing the expected squared prediction error when $Y$ is real-valued. In high-dimensional problems such as deep neural network models, the optimal solution may have very large number of parameters, leading to overfitting. That is, the optimal solution $f_{\hat{\pmb{\bm{\beta}}}}(X)$ may predict well in the training dataset that was used to fit the model but may underperform when used for prediction on a new `test' dataset. One approach that is commonly employed to reduce overfitting is to add a penalty function $\mathbb{P}_{\lambda, \kappa}(\pmb{\bm{\beta}}) = \lambda \mathbb{P}_{\kappa}(\pmb{\bm{\beta}})$ to the loss function, where $\lambda$ and $\kappa$ are tuning parameters. Thus the new overall objective function of interest is $\mathcal{L}(Y, f_{\pmb{\bm{\beta}}}(X)) + \lambda \mathbb{P}_{\kappa}(\pmb{\bm{\beta}})$ and the optimization program is \begin{equation}\pmb{\bm{\hat{\beta}}} = \argminB_{\pmb{\bm{\beta}}} \left[\mathcal{L}(Y, f_{\pmb{\bm{\beta}}}(X)) + \lambda \mathbb{P}_{\kappa}(\pmb{\bm{\beta}}) \right]. \end{equation}

 Various penalty functions have been proposed in the literature. One way to classify the currently existing penalties are based on their convexity: that is, convex versus nonconvex penalties. Convex penalties are advantageous from an optimization perspective but could lead to biased results. Employing nonconvex penalties, on the other hand, is not so ideal for optimization but it could yield unbiased or nearly unbiased parameter values especially for large parameter values. Another way to classify the penalties is based on whether they are singular at origin or not. Penalties which are singular at origin, set many elements of the solution vector $\pmb{\bm{\hat{\beta}}}$  to zero, and thus could be used for variable selection. Penalties which are smooth at origin, on the other hand, cannot be used for variable selection. However, in many large dimensional statistical problems such as deep neural networks, variable selection is not of any concern; the primary concern in such problems is reducing the prediction error. The nonconvex penalty functions that have appeared in the literature so far, are all singular at origin, to the best of our knowledge. In this paper, we introduce a novel nonconvex penalty function which is smooth at origin: $\mathbb{P}_{\kappa}(\pmb{\bm{\beta}}) = \sum_{j=1}^{p}\mathcal{P}_{\kappa}(\beta_{j})$ for $\pmb{\bm{\beta}} = (\beta_{1}, \ldots, \beta_{p})^{T} \in \mathbb{R}^{p}$ where
\begin{equation} \mathcal{P}_{\kappa}(\beta) = 1 - e^{-\kappa \beta^{2}},\;\mathrm{for}\; \beta \in \mathbb{R}. \end{equation} In the following sections of the paper the new penalty function introduced in Eq. (2) will be labelled as the Gaussian penalty function. Figure 1 below plots the new penalty function, and also one example penalty function from each of the following categories: (a) convex, singular-at-origin (Lasso), (b) convex, smooth-at-origin (Ridge), (c) nonconvex, singular-at-origin (arctan), and (d) nonconvex, smooth-at-origin (Gaussian).

\begin{figure}[H]
  \begin{center}
  \hspace*{-1.0cm}
  \includegraphics[height=3.5in,width=7in,angle=0]{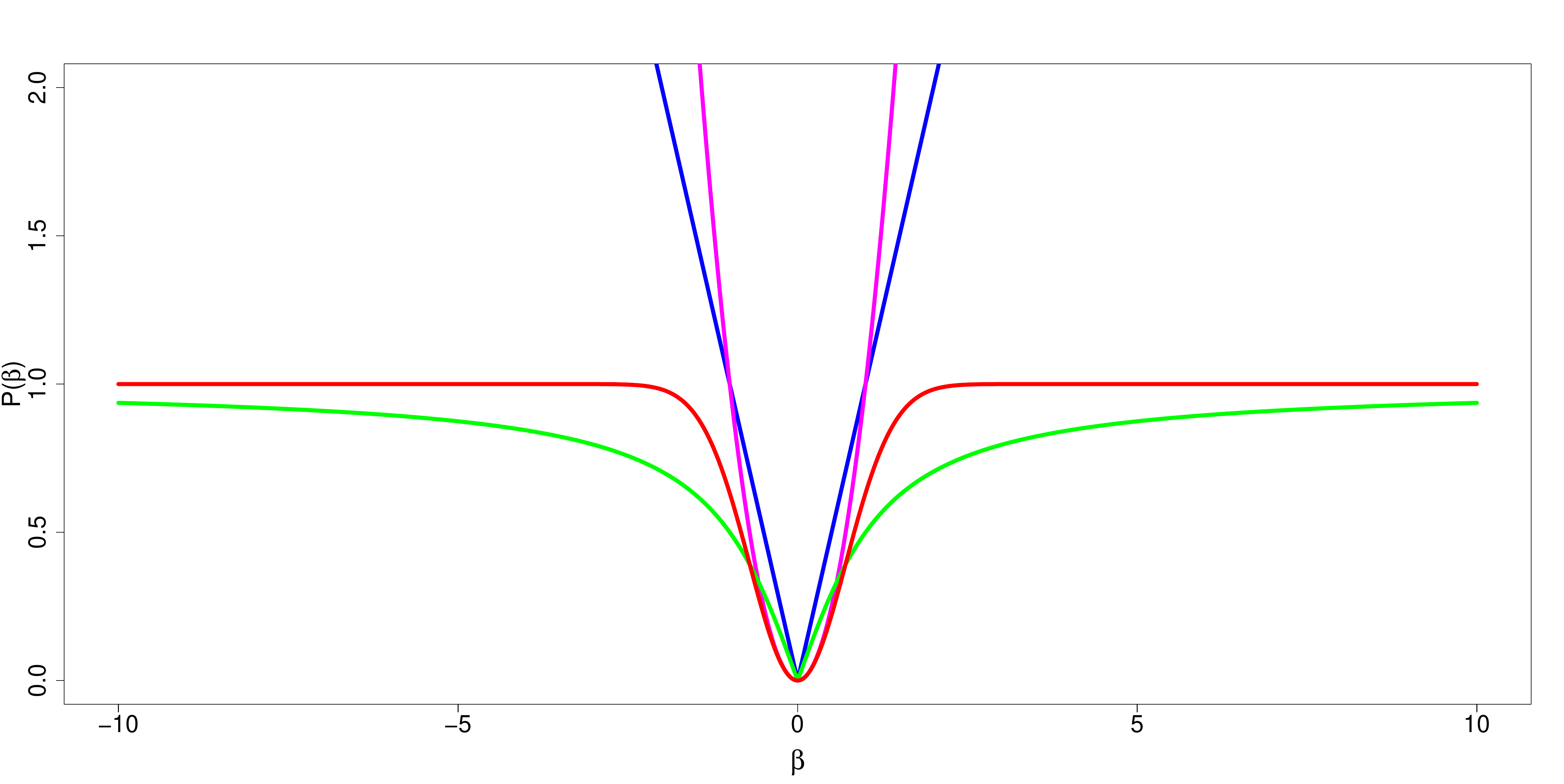} 
  \caption{Various examples of penalty functions: Lasso, $\mathcal{P}$($\beta$) = $\abs{\beta}$ (blue); Ridge, $\mathcal{P}$($\beta$) = $\beta^{2}$ (magenta); arctan, $\mathcal{P}$($\beta$) = $\frac{2}{\pi}\arctan (\abs{\beta})$ (green); and Gaussian, $\mathcal{P}$($\beta$) = $1- e^{-\beta^{2}}$ (red). Note that Ridge and Gaussian penalties overlap in a small neighborhood near $\beta =$ 0.}
  \end{center}
\end{figure}


\textbf{Remark 1.} For any real numbers $x, y$ (say, $x < y$ without loss of generality), \begin{align} |\mathcal{P}_{\kappa}(x) - \mathcal{P}_{\kappa}(y)| &= |\mathcal{P}_{\kappa}'(u)||x-y|\;\mathrm{for}\;u \in (x,y) \nonumber \\ &\leq (\sqrt{2\kappa}e^{-1/2})|x-y|, \nonumber  \end{align} since $\mathcal{P}_{\kappa}'(u) = 2\kappa u e^{-\kappa u^{2}} \leq \sqrt{2\kappa}e^{-1/2}$. Hence $\mathcal{P}_{\kappa}(\cdot)$ is Lipschitz continuous.

\section{Asymptotic Consistency Results for Ordinary Least Squares Linear Regression}

In this section, we focus on the model $f_{\pmb{\bm{\beta}}}(X) = X^{T}\pmb{\bm{\beta}}$ and the loss function $\mathcal{L}(Y, f_{\pmb{\bm{\beta}}}(X)) = (Y - f_{\pmb{\bm{\beta}}}(X))^{2}$ with real-valued $Y$; that is, ordinary least squares (OLS) linear regression. We prove two asymptotic results, similar to the theorems in Knight and Fu, 2000. The first proposition is a consistency result and the second proposition is a $\sqrt{n}$-consistency result, which would help us to better understand the bias involved with the regularized estimator $\pmb{\bm{\hat{\beta}}}$ for OLS with Gaussian penalization. Let $\bm{x_{i}}^{T}, y_{i}, i = 1, \ldots, n$, denote $n$ realizations of the random vector $X$ and the random variable $Y$, respectively, and let $\bm{X} = $[$\bm{x_{1}}^{T}, \ldots, \bm{x_{n}}^{T}$]$^{T}$ denote the corresponding design matrix. Thus the empirical linear regression model that we are interested in is \begin{equation} \bm{y} = \bm{X}^{T}\pmb{\bm{\beta}} + \bm{e}, \label{OLS} \end{equation} where $\bm{y} = (y_{1}, \ldots, y_{n})^{T}$ and $\bm{e}$ is a vector of i.i.d. random variables with mean zero and variance $\sigma^{2}$. The empirical loss function corresponding to this model is $\mathcal{L}_{n}(\bm{y}, \bm{X}^{T}\pmb{\bm{\beta}}) = \sum_{i=1}^{n}(y_{i} - \bm{x_{i}}^{T}\pmb{\bm{\beta}})^{2}.$ Without loss of generality, we assume that the covariates in Eq. (\ref{OLS}) are centered and the mean of the $y_{i}$'s is equal to zero.

\noindent \textbf{\underline{Proposition-1}}: With $\bm{x_{i}}^{T}$ denoting the $i^{th}$ row of $\bm{X}^{T}$, we assume\[ C_{n} = \frac{1}{n} \sum_{i=1}^{n} \bm{x_{i}}\bm{x_{i}}^{T} \rightarrow C, \] where $C$ is a nonnegative definite matrix. We also assume that \[ \frac{1}{n} \max_{1 \leq i \leq n} \bm{x_{i}}^{T}\bm{x_{i}} \rightarrow 0, \] $C$ is non-singular, $\kappa > 0$, the components $e_{i}$'s of the error vector in Eq. (\ref{OLS}) are i.i.d with mean zero and variance $\sigma^{2}$.  If $\lambda_{n}/n \rightarrow \lambda_{0}\; (\lambda_{n}, \lambda_{0} \geq 0)$ then $\bm{\hat{\beta}}_{n} \rightarrow_{p} \argminB (Z)$ where \[ Z(\bm{\phi}) = (\bm{\phi} - \bm{\beta})^{T}C(\bm{\phi} - \bm{\beta}) + \lambda_{0} \sum_{j=1}^{p}\left(1 - \exp\{-\kappa\phi_{j}^{2}\} \right). \] Since $\argminB (Z) = \bm{\beta}$, $\bm{\hat{\beta}}_{n}$ is consistent provided $\lambda_{n} = o(n)$.

\textbf{Proof:}
    We prove Proposition-1 by adapting the proof of Theorem 1 in Knight and Fu, 2000. We have $\bm{\hat{\beta}}_{n} = \argminB (Z_{n})$ where \[ Z_{n}(\bm{\phi}) = \frac{1}{n}\sum_{i=1}^{n} (y_{i} - \bm{x}_{i}^{T}\bm{\phi})^{2} + \frac{\lambda_{n}}{n} \sum_{j=1}^{p}\left(1 - \exp\{-\kappa\phi_{j}^{2}\} \right). \] We rewrite \begin{align*} Z_{n}(\bm{\phi}) &= \frac{1}{n}\sum_{i=1}^{n} \left[ (y_{i} - \bm{x}_{i}^{T}\bm{\beta}) - \bm{x}_{i}^{T}(\bm{\phi} - \bm{\beta})\right]^{2} + \frac{\lambda_{n}}{n} \sum_{j=1}^{p}\left(1 - \exp\{-\kappa\phi_{j}^{2}\} \right). \end{align*} Since $\mathbb{E}(e_{i}) = 0,\; \mathbb{E}(e_{i}^{2}) = \sigma^{2},\; C_{n} \rightarrow C \;\mathrm{and}\; \lambda_{n}/n \rightarrow \lambda_{0}$, \begin{align*} \mathbb{E}(Z_{n}(\phi)) &= \sigma^{2} + (\bm{\phi} - \bm{\beta})^{T}C_{n}(\bm{\phi} - \bm{\beta}) + \frac{\lambda_{n}}{n} \sum_{j=1}^{p}\left(1 - \exp\{-\kappa\phi_{j}^{2}\} \right)\\ &\rightarrow \sigma^{2} + Z(\bm{\phi}), \;\; \mathrm{as\;} n \rightarrow \infty, \end{align*} which implies pointwise convergence in probability. Next we show \begin{equation} \sup_{\phi \in K} \abs{Z_{n}(\phi) - Z(\phi) - \sigma^{2}} \rightarrow_{p} 0, \;\; \mathrm{for \; any\; compact\; set\;} K. \label{Prop1.eq1}\end{equation} Utilizing stochastic Ascoli lemma we have to show \textit{a.s.} equicontinuity for $Z_{n}$ in order to prove (\ref{Prop1.eq1}). A sufficient condition for stochastic equicontinuity is \textit{a.s.} Lipschitz continuity for $Z_{n}$. Since the OLS loss function is convex (and hence Lipschitz continuous over a compact set $K$) and since we have established Lipschitz property for the Gaussian penalty function in Remark 1, the uniform convergence claim in (\ref{Prop1.eq1}) follows. Since we assume $\lambda_{n} \geq 0$ and $\kappa > 0$, we have \[ Z_{n}(\phi) \geq \frac{1}{n} \sum_{i=1}^{n} \left( y_{i} - \bm{x}_{i}^{T}\bm{\phi} \right)^{2} = Z_{n}^{(0)}(\bm{\phi}), \;\; \mathrm{for\;all}\; \bm{\phi}. \] Since $\argminB (Z_{n}^{(0)}) = O_{p}(1)$, we get \begin{equation} \bm{\hat{\beta}}_{n} = O_{p}(1). \label{Prop1.eq2} \end{equation} Equations (\ref{Prop1.eq1}) and (\ref{Prop1.eq2}) imply that $\bm{\hat{\beta}}_{n}$ is consistent. This completes the proof of Proposition-1.

We now state and prove a $\sqrt{n}$-consistency result for the Gaussian penalized OLS estimators using proof methods similar to those in Theorem 3 in Knight and Fu, 2000.

\noindent \textbf{\underline{Proposition-2}}: If $\lambda_{n}/\sqrt{n} \downarrow \lambda_{0} \geq 0$, then $\sqrt{n}(\hat{\bm{\beta}}_{n} - \bm{\beta}) \rightarrow_{d} \mathrm{argmin}(V)$, where
\[ V(\mathbf{u}) = -2\mathbf{u}^{T}W + \mathbf{u}^{T}C\mathbf{u} + 2\lambda_{0}\kappa\sum_{j=1}^{p}\left[u_{j}\beta_{j}\exp\{-\kappa \beta_{j}^{2} \} \right] \] and $W$ has a $N(\mathbf{0}, \sigma^{2}C)$ distribution.

    \textbf{Proof:} Consider \begin{align} V_{n}(\mathbf{u}) &= \sum_{i=1}^{n}\left[(e_{i} - \mathbf{u}^{T}\bm{x}_{i}/\sqrt{n})^{2} - e_{i}^{2} \right] \nonumber \\ & \;\;\;\;\;\;\;\;\;\;\;+ \lambda_{n}\sum_{j=1}^{p}\left[(1-\exp \{-\kappa[\beta_{j} + u_{j}/\sqrt{n}]^{2}\}) - (1-\exp\{-\kappa \beta_{j}^{2}\}) \right]. \nonumber \end{align}

 Using Tayor series expansion,
\begin{align}  q_{n}(u_{j}) \equiv \lambda_{n}&\left[\exp\{-\kappa \beta_{j}^{2}\} - \exp \{-\kappa[\beta_{j} + u_{j}/\sqrt{n}]^{2}\} \right]  \nonumber \\ &=  2\lambda_{n}\kappa \beta_{j}\exp\{-\kappa \beta_{j}^{2} \}u_{j}/\sqrt{n} + o(1) \nonumber \\ &\rightarrow 2\lambda_{0}\kappa u_{j}\beta_{j}\exp\{-\kappa \beta_{j}^{2} \},\;\mathrm{since}\;\lambda_{n}/\sqrt{n} \rightarrow \lambda_{0}\; \mathrm{by\; assumption}. \label{Prop2.eq1}  \end{align}

 The above pointwise convergence in (\ref{Prop2.eq1}), continuity of $q_{n}$ and the fact that $q_{n} \geq q_{n+1}$ for large $n$ implies that convergence is uniform over $\mathbf{u}$ in compact sets (see e.g. Theorem 7.13 in. p.150 in Rudin, 1976). It follows that \[ V_{n}(\cdot) \rightarrow_{d} V(\cdot)\] on the space of functions topologized by uniform convergence on compact sets. Following Knight and Fu, 2000, and Kim and Pollard, 1990, in order to prove that $\mathrm{argmin}(V_{n}) \rightarrow_{d} \mathrm{argmin}(V)$, it suffices to show that $\mathrm{argmin}(V_{n}) = O_{p}(1)$. Based on Markov's lemma, in turn, it suffices to show that $\mathbb{E}(V_{n})$ and $\mathrm{Var}(V_{n})$ are both $O(1)$ (see e.g. remark at the end of Example 2.6 in van der Vaart, 1998, p.10).
\begin{align} \mathbb{E}(V_{n}) &= \frac{1}{n}\sum_{i=1}^{n}\mathbf{u}^{T}(\bm{x}_{i}\bm{x}_{i}^{T})\mathbf{u}
- \lambda_{n}\sum_{j=1}^{p}\left[e^{-\kappa[\beta_{j} + u_{j}/\sqrt{n}]^{2}} - \varepsilon^{-\kappa \beta_{j}^{2}} \right] \nonumber \\ &\rightarrow \mathbb{E}(V), \;\mathrm{a\,constant.}\;\; (\mathrm{i.e.}\; \mathbb{E}(V_{n}) = O(1)). \nonumber \\ \mathrm{Var}(V_{n}) &= \frac{4}{n}\sum_{i=1}^{n}\mathbf{u}^{T}(\bm{x}_{i}\bm{x}_{i}^{T})\mathbf{u} \nonumber \\ &\rightarrow 4\mathbf{u}^{T}C\mathbf{u} = \mathrm{Var}(V)\;\; (\mathrm{i.e.}\; \mathrm{Var}(V_{n}) = O(1)). \nonumber \end{align}  This completes the proof of Proposition-2.

\textbf{Remark 2.} Proposition 2 essentially states that the regularized estimator $\pmb{\bm{\hat{\beta}}}$ for OLS with Gaussian penalization is asymptotically biased. However it is interesting to compare the bias terms of the form $u\beta\exp\{-\kappa \beta^{2}\}$ in the statement of Proposition 2 with the bias terms for Ridge estimators, $u\,sgn(\beta)\beta^{2}$, in Theorem 2 in Knight and Fu, 2000. Based on these terms, it is easy to observe that the bias for Gaussian penalty based OLS estimator goes to zero exponentially fast as $\beta \rightarrow \infty$ while as bias for the Ridge estimator increases quadratically.

\textbf{Remark 3. (Orthonormal design case)} Further understanding, especially using visual means, may be obtained about the OLS estimator based on the Gaussian penalty by considering the orthonormal design case: $\bm{X}^{T}\bm{X} = I$. In this case, the $j^{th}$ element $\hat{\beta}_{j}$ of $\pmb{\bm{\hat{\beta}}}$ satisfies (H\"{a}rdle and Simar, 2005, Chapter 7), \begin{equation} \hat{\beta}_{j} = \argminB_{\beta} \left[ -2\hat{\beta}_{j}^{O}\beta + \beta^{2} + \lambda (1-e^{-\kappa \beta^{2}}) \right], \label{orthoobj}\end{equation} where $\hat{\beta}_{j}^{O}$ is the $j^{th}$ element of $\hat{\pmb{\bm{\beta}}}^{O} = \bm{X}^{T}\bm{y}$.

Figure 2 below exhibits an animation movie of the objective function in Eq. (\ref{orthoobj}) for a grid of $\lambda$ values. It is easily observed that as $\lambda$ increases two local minima appears. For smaller values of $\lambda$ the argument of the global minimum correspond to the unbiased OLS estimate without regularization. As $\lambda$ increases, the global minimum shifts to the other local minimum for which the argument is a value very close to zero. In addition, we see from the animation movie that objective function is smooth at both local minima.

\begin{figure}[H]
\begin{center}
\animategraphics[label=lambdai, height=3in, width=3.35in, controls, timeline=timeline.txt]{4}{lambdai.}{1}{16}\caption{Animation movie showing two minima, one at 0 and the other at $\hat{\beta}_{0}$ for the objective function $-2\hat{\beta}_{0}\beta + \beta^{2} + \lambda (1-e^{-\kappa \beta^{2}})$ with $\hat{\beta}_{0} = 3$, as $\lambda$ increases from 0.1 to 15.1 with a step-size of 1.0. We took for $\kappa = 10$ for this plot. A green vertical line through $\beta = \hat{\beta}_{0}$ is plotted for reference.}
\mediabutton[
jsaction={
if(anim[’lambdai’].isPlaying)
anim[’lambdai’].pause();
else
anim[’lambdai’].playFwd();
}
]{\fbox{Play/Pause}}
\end{center}
\end{figure}

\section{Case studies: Empirical analyses with deep neural network and convolutional neural network models}

 In this section we empirically evaluate the performance of the new penalty in deep neural network setting for which the loss function is known to be nonconvex (Bishop, 2006). We assessed the performance of regularized deep neural networks with the Gaussian penalty function, by applying them on seven datasets (MNIST, FMNIST, RCV1, CIFAR-10, CIFAR-100, SVHN and ImageNet). Convolutional neural network architecture was employed on four of the seven datasets: CIFAR-10, CIFAR-100, SVHN and ImageNet. Specifics of the DNN and CNN analyses and architectures, and details of all the seven datasets are given below. The architectures and the datasets are the same as ones used in our previous paper, Vettam and John, 2022. For completeness, we describe them again below.

  \subsection*{Deep Neural Network analyses}

  A deep neural network with 5 layers was utilized on MNIST, FMNIST and RCV1 datasets. For MNIST and FMNIST datasets, the DNN model had 1024 nodes within each layers; for RCV1 dataset 512 nodes were utilized within each layer. The optimal weights of the fitted deep neural networks were estimated by minimizing the total cross entropy loss function. We used batch gradient descent algorithm with early stopping. To avoid the vanishing/exploding gradients problem, the weights were initialized to values obtained from a normal distribution with mean zero and variance $4/(n_{i}+n_{(i-1)})$ where $n_{i}$ is the number of neurons in the $i^{th}$ layer (Glorot and Bengio, 2010; He, Zhang, Ren and Sun, 2015). Rectified linear units (ReLU) function was used as the activation function.

  The training data was randomly split into multiple batches. During each epoch, the gradient descent algorithm was sequentially applied to each of these batches resulting in new weights estimates. At the end of each epoch, the total validation loss was calculated using the validation set. When twenty consecutive epochs failed to improve the total validation loss, the iteration was stopped. The maximum number of epochs was set at 250. The weights estimate that resulted in the lowest total validation loss was selected as the final estimate. Since there was a random aspect to the way the training sets were split into batches, the whole process was repeated three times with seed values 1, 2, and 3. The reported test error rates are the median of the three test error rates obtained using each of these seed values.

\begin{figure}[H]
\begin{center}
\hspace*{1.0cm}
\includegraphics[height=2.25in,width=3in,angle=0]{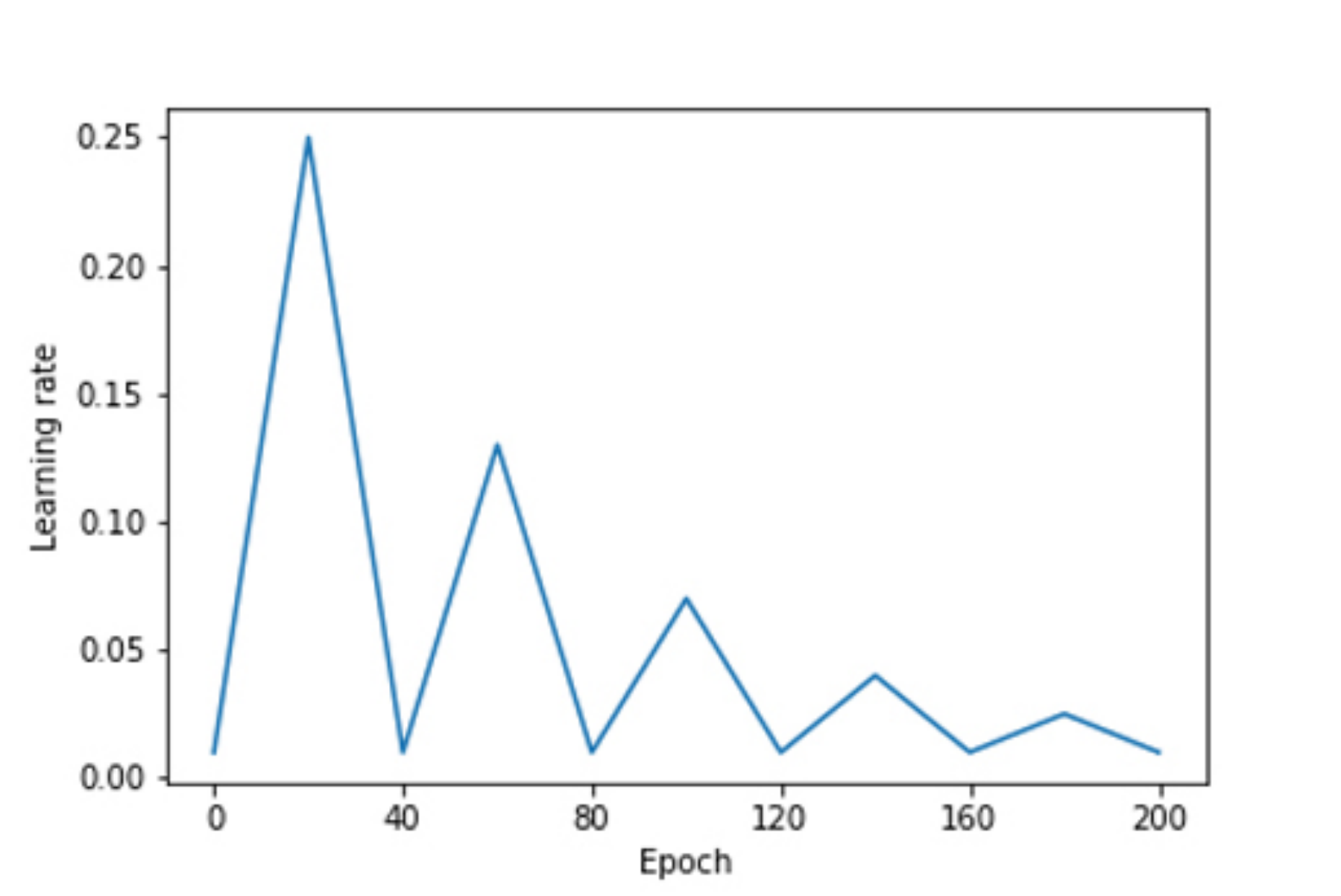}
\caption{Learning rate plot}
\end{center}
\end{figure}

   A triangular learning rate schedule was used because it produced the lowest test error rates (Smith, 2017). The learning rates varied from a minimum of 0.01 to a maximum of 0.25 (see Figure 3). For all penalty functions the optimal $\lambda$ was found by fitting models with logarithmically equidistant values in a grid. We used Python ver. 3.6.7rc2 and TensorFlow ver.1.12.0 for the calculations. Details of the three datasets used for DNN analyses (without the CNN architecture) are as follows. A general overview of the datasets and the DNN model specifications is given in Table 1 further below.

\underline{MNIST}:

Modified National Institute of Standards and Technology (MNIST) dataset is a widely used toy dataset of 60,000 grey-scale images of hand-written digits. Each image has $28 \times 28 = 784$ pixels. The intensity measures of these 784 pixels form the input variables of the model. The dataset was split into 48,000 training set, 2000 validation set, and 10,000 test set.

\underline{FMNIST}:

Fashion MNIST dataset consists of 60,000 grey-scale images of various types of clothing such as shirts, pants, and caps. There are 10 classes in total. The images have $28 \times 28 = 784$ pixels whose intensity measures were used as the input variables of the model. The 60,000 images were split into 45,000 training set, 5000 validation set, and 10,000 test set. FMNIST is very similar to MNIST because of similar characteristics except that the FMNIST test error rates tend to be much higher than MNIST test error rates.

\underline{RCV1}:

Reuters Corpus Volume I (RCV1) is a collection of 804,414 newswire articles labelled as belonging to one or more of 103 news categories (Lewis, Yang, Rose, Li, 2004). In our analysis, only single-labelled data points were used and all the multi-labelled data points were excluded resulting in a dataset consisting of news wire articles from 50 news categories. The cosine-normalized, log TF-IDF values of 47,236 words appearing in these news articles were used as the input variables for the model. The training set consisted of 13,355 data points, the validation set consisted of 2000 data points and the test set consisted of 49,565 data points.

\scriptsize
\begin{center}
    \begin{tabular}{ | l || c | c | c | c | c | c | c | }
    \hline
    \multicolumn{8}{|c|}{Table 1. Overview of dataset and DNN model specifications}   \\ \hline
    Dataset & Domain & Dimensionality & Classes	& DNN & 	Training&	Validation & Test   \\
        & & & 	& Specifications & 	Set & Set &	Set  \\ \hline \hline
    MNIST   & Visual & 784 ($28 \times 28$  & 10 & 5 layers,	& 48000	& 2000	& 10000 \\
            &        &  greyscale) &  & 1024 nodes & & &  \\ \hline
    FMNIST  & Visual & 784 ($28 \times 28$  & 10 & 5 layers,	& 45000	& 5000	& 10000 \\
            &        &  greyscale) &  & 1024 nodes & & &  \\ \hline

    Reuters	& Text	& 47236 & 	50	& 5 layers,
	& 13355	& 2000	& 49565 \\
            & 	& &	& 512 nodes	& &	&  \\ \hline
    \end{tabular}
\end{center}
\normalsize

  \subsection*{Convolutional Neural Network analyses}

The CNN architecture that we used consisted of three convolutional ``blocks'' followed by a hidden ``block''. Each of the three convolutional blocks consisted of a convolutional layer with 96, 128 and 256 filters respectively, kernel size of 5, stride value of 1, and “same” padding, followed by batch normalization with momentum value for the moving average set to 0.9. This was further followed by a ReLU activation layer and finally a max-pooling layer with kernel-size of 3, stride value of 2, and ``same'' padding. In the hidden block, the data was first flattened and then passed through two fully connected hidden layers of 4096 nodes with ReLU activation function. Finally, the signals were classified into various categories by calculating the sparse softmax cross entropy values between logits and labels. The CNN model specification is further described in Figure 4.

\begin{figure}[H]
\begin{center}
\hspace*{-1.2cm}
\includegraphics[height=1.8in,width=5.4in,angle=0]{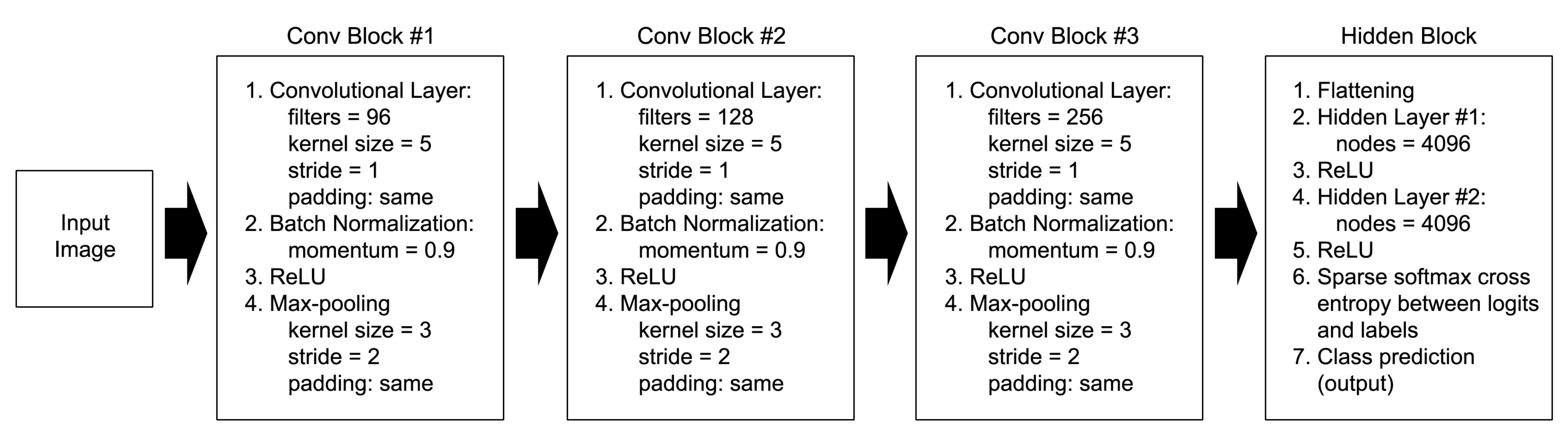}
\caption{Specifications for Convolutional Neural Network architecture}
\end{center}
\end{figure}

We assessed the performance of regularized CNNs with the Gaussian penalty function by applying them on four datasets (CIFAR10, CIFAR100, SVHN, ImageNet). More details of the four datasets used for CNN analyses are given below. These details and CNN specifications are summarized in Table 2.

\underline{CIFAR-10 and CIFAR-100}:

CIFAR10 and CIFAR100 are datasets of 50000 32 $\times$ 32 color training images and 10000 test images. The images are classified into 10 categories for CIFAR-10 and 100 categories for CIFAR-100.

\underline{Street View House Numbers (SVHN)}:

The SVHN dataset is a collection of 32 $\times$ 32 color images of digits 0 to 9 obtained from house numbers in Google Street View images. SVHN is a bigger dataset than MNIST: 71257 training, 2000 validation, 26032 test.The images are in a natural setting unlike in MNIST, and hence the classification problem is a bit harder.

\underline{ImageNet}:

The original ImageNet dataset for ILSVRC-2014 challenge had a total of 476688 color images in the training and validation sets (http://www.image-net.org/challenges/LSVRC/2014/). These images were of varying sizes. Since our analysis is mainly for illustration purposes, we used a subset of the full ILSVRC-2014 dataset with 200 classes. The color images in this subset were resized from their original sizes to 64 $\times$ 64 pixels. The smaller subset helped us to reduce substantially the computational time and resources required to train the models. However, the subset that we used is still large and challenging enough to obtain meaningful comparisons. In a sense, reducing the image counts and resolution made the classification problem even more challenging in terms of accuracy compared to the original ILSVRC-2014 classification problem. For each class, there were 450 training images, 50 validation images, and 50 test images. The final image counts for training, validation, and test datasets in our subset were 90000, 10000 and 10000, respectively. The class IDs of the images that we used are available upon request as a text file.

\scriptsize
\begin{center}
    \begin{tabular}{ | l || c | c | c | c | c | c | }
    \hline
    \multicolumn{7}{|c|}{Table 2. Overview of datasets used for Convolutional Neural Networks}   \\ \hline
    Dataset & Domain & Dimensionality & Classes	& 	Training &	Validation & Test   \\
        & & & 	& Set & Set &	Set  \\ \hline \hline
    CIFAR-10  & Visual & 3072 ($32 \times 32 \times 3$  & 10 & 48000	& 2000	& 10000 \\
            &        &  color) &  & & & \\ \hline
    CIFAR-100  & Visual & 3072 ($32 \times 32 \times 3$  & 100 & 48000	& 2000	& 10000 \\
            &        &  color) &  & & &  \\ \hline
    SVHN & Visual & 3072 ($32 \times 32 \times 3$  & 10 & 71257	& 2000	& 26032 \\
            &        &  color) &  & & & \\ \hline
    Reduced size& Visual & 12288 ($64 \times 64 \times 3$  & 200 & 90000 & 10000	& 10000  \\
    ImageNet&        &  color) &  & &  &\\ \hline
    \end{tabular}
\end{center}
\normalsize

\subsection{Results}

For nonconvex penalties there is an extra parameter that we have to take into consideration: $a$ for SCAD, $b$ for MCP, $\varepsilon$ for Laplace, $\gamma$ for arctan and $\kappa$ for Gaussian. In previous literature (Fan and Li, 2001), $a = 3.7$ has been suggested as an optimal parameter for SCAD based on Bayes risk criterion. Under different settings, $b  = 1.5, 5$ and $20$ have been considered as optimal values for the MCP parameter (Breheny and Huang, 2011). $a = 3.7$ for SCAD and $b  = 1.5, 5$ and $20$ are the parameter values that we considered for all DNN analyses in this paper. After trial and error runs with one dataset (MNIST), we picked $\gamma =$ 1 and 100 for arctan and $\varepsilon = 10^{-7}$ for Laplace penalty (same as the parameters used in Vettam and John, 2022). Based on trial and error again, $\kappa =$ 10 was chosen for the Gaussian penalty. Since SCAD and MCP did not perform as well as the other penalties in the DNN analyses, we restricted our focus to only Laplace, arctan and Gaussian as nonconvex penalty-candidates for CNN analyses.

\scriptsize
\begin{center}
    \begin{tabular}{ | l || c | c | c |c | c | c | c |  }
    \hline
    \multicolumn{8}{|c|}{Table 3. Median test error rates at optimal $\lambda$  } \\ \hline
    Penalty function & \multicolumn{3}{|c|}{Datasets, DNN} & \multicolumn{4}{|c|}{Datasets, CNN} \\ \hline
            & MNIST & FMNIST & RCV1 & CIFAR-10 & CIFAR-100 & SVHN & ImageNet \\ \hline\hline
    None & 1.87 (0.11) & 11.94 (0.13) & 14.66 (0.32) & 19.58 (0.24) & 50.64 (0.25) & 6.63 (0.10) & 42.85 (0.19)\\ \hline
   $\mathrm{L}_{1}$ (Lasso) & 1.24 (0.05) & 10.06 (0.21) & 12.97 (0.15)& 14.71 (0.15)  & 43.22 (0.36) & 5.32 (0.06) & 41.32 (0.42) \\ \hline
   $\mathrm{L}_{2}$ (Ridge) & 1.23   (0.01) & 10.15 (0.02) & 13.77 (0.33)& 14.32  (0.25) &  41.11 (0.61)  & 5.29 (0.11) & 41.64 (0.28) \\ \hline
     SCAD (a = 3.7) & 1.80 (0.07) & 11.45 (0.23) & 13.96 (0.03)& 18.56 (0.42) & 49.50 (0.39) & 6.51 (0.04) & 42.12 (0.20) \\ \hline
      MCP (b = 1.5) & 1.60 (0.22) & 11.39 (0.15) & 13.33 (0.22)& x & x & x & x\\ \hline
      MCP (b = 5)   & 1.67 (0.21) & 11.39 (0.12) & 14.44 (0.45)& 18.47 (0.49) & 50.22 (0.52) & 6.33 (0.15) & 41.99 (0.30) \\ \hline
      MCP (b = 20)  & 1.65 (0.16) & 11.33 (0.04) & 14.36 (0.37)& x & x & x & x \\ \hline
      Laplace ($\varepsilon = 10^{-7}$)  & 1.23  (0.05) & 9.98 (0.25) & \color{blue}12.94 \color{black} (0.21)& 18.87 (0.36) & 49.79 (0.69) & 6.43 (0.14) & 42.22 (0.15) \\ \hline
      arctan ($\gamma$ = 1)    & 1.26 (0.04)& 9.87 (0.13) & 13.41 (0.12)& 14.61 (0.06) & 43.58 (0.04) & 5.37 (0.05) & 40.98  (0.20) \\ \hline
      arctan ($\gamma$ = 100)  & 1.25 (0.03) & 9.93 (0.23) & 13.81 (0.32)& 14.39 (0.17) & 43.33 (0.26) & \color{blue} 5.19 \color{black} (0.09)   & 42.61 (0.09) \\ \hline
       Gaussian ($\kappa$ = 10)  & \color{blue} 1.19 \color{black} (0.03) & \color{blue} 9.72 \color{black} (0.17) & 13.41 (0.31)& \color{blue} 13.99 \color{black} (0.15) & \color{blue} 39.87 \color{black} (0.20) & 5.48 (0.17)   & \color{blue} 40.30 \color{black} (0.90) \\ \hline
      \multicolumn{8}{|c|}{Standard errors are given in parentheses} \\ \hline
    \end{tabular}
\end{center}
\normalsize

Results are summarized in Table 3. All the rows in Table 3 except for the last row is exactly the same as in Vettam and John, 2022. Only the last row (that is, the one corresponding to the Gaussian penalty) is the new addition. We present all the other rows for completeness and for convenience. The best error rates for each dataset is highlighted in blue. From Table 3, it is seen that the Gaussian penalty yielded the best error rates in 5 out of the 7 datasets considered: specifically, for MNIST, FMNIST, CIFAR-10, CIFAR-100 and ImageNet. For RCV1, the best error rate was obtained by the Laplace penalty and for SVHN, the best error rate was obtained by the arctan penalty ($\gamma =$ 100); that is, for these two datasets the Gaussian penalty did not improve the accuracy rates reported in Vettam and John, 2022. It is interesting to note that for all the seven datasets, the best accuracy was obtained by a nonconvex penalty. Also, interesting to point out that for CIFAR-10 and CIFAR-100, the second best accuracy was obtained with $L_{2}$ (Ridge) penalty. In other words, for these two datasets, the best accuracy rate seen in the previous paper, Vettam and John, 2022, was for $L_{2}$. However, the newly introduced Gaussian penalty function which has similarities to $L_{2}$ near the origin but nonconvex unlike $L_{2}$, was able to lower the error rate.

\section*{Conclusions and brief discussion}

In this paper, we introduced a novel nonconvex penalty function for regularization in high-dimensional statistical learning algorithms. The novel function, which we labelled as `Gaussian penalty function' is smooth at origin unlike other nonconvex penalty functions such as SCAD, MCP, Laplace and arctan.  We studied asymptotic properties of the regularized estimators under the ordinary least squares (OLS) loss function. Asymptotic results showed bias for the regularized OLS estimator. However, this bias vanished exponentially fast as the value of the underlying OLS parameter increased.

We also studied empirically the performance of the new regularizer in deep learning setting by applying it on seven different datasets. Deep neural net (DNN) architecture was employed on three datasets, and convolutional neural net (CNN) architecture was employed on four datasets. The new regularization method exhibited the best performance in five out of the seven datasets (two out of three which employed the DNN architecture and three out of four which employed CNN architecture). The fact that the new penalty did not yield the best results for two of the datasets, suggests scope for improvement. One direction for future work would be to tune over all the penalty functions (or may be the nonconvex penalty functions separately from the convex penalty functions) for any given dataset. Another direction will be to introduce randomness into the hyperparameter selection of the nonconvex penalty functions, especially for deep neural network architectures. Specifically, the idea will be to randomly select an hyperparameter for each iteration in the training phase of the neural net, effectively varying the objective function randomly during the learning phase.

\section*{Codes availability} Sample python codes used for the CNN analysis in the paper are posted on the following github page: `github.com/mjohn5/dnn\_gaussian\_regularization'.

\section*{Acknowledgements} The first two authors are grateful to the third author for suggesting the new penalty function (i.e. the Gaussian penalty) presented in this paper. 

\section*{References}

Bishop, C.M. (2006). Pattern Recognition and Machine Learning. Springer. New York.

Breheny, P., Huang, J. (2011). Coordinate descent algorithms for nonconvex penalized regression, with applications to biological feature selection. The Annals of Applied Statistics, 5(1): 232-253.

Fan, J., Li, R. (2001). Variable selection via nonconcave penalized likelihood and its oracle properties. Journal of the American Statistical Association, 96: 1348–1360.

Frank, I., Friedman, J. (1993). A statistical view of some chemometrics regression tools. Technometrics, 35: 109–148.

Fu, W. J. (1998). Penalized regressions: the Bridge versus the Lasso. J. Comput. Graph. Statist, 7: 397–416.

Glorot, X., and Bengio, Y. (2010). Understanding the difficulty of training deep feedforward neural networks. PMLR 9: 249-256.

H\"{a}rdle, W.K., Simar, L. (2019). Applied Multivariate Statistical Analysis. $5^{th}$ edition. Springer Verlag, Berlin-Heidelberg-New York.

Hastie, T., Tibshirani, R., Friedman, J. H. (2001). The elements of statistical learning: Data mining, inference, and prediction. New York: Springer.

He, K., Zhang, X., Ren, S., Sun, J. (2015). Delving Deep into Rectifiers: Surpassing Human-Level Performance on ImageNet Classification. arXiv eprint 1502.01852.

Hoerl, A.E., Kennard, R. (1970). Ridge regression: Biased estimation for nonorthogonal problems. Technometrics, 12: 55-67.

Kim, J., Pollard, D. (1990).  Cube root asymptotics. The Annals of Statistics. 18: 191–219.

Knight, K., Fu, W. (2000). Asymptotics for lasso-type estimators. Ann. Statist. 28(5): 1356-1378.

Lewis, D. D., Yang, Y., Rose, T. G.,  Li, F. (2004). RCV1: A new benchmark collection for text categorization research. The Journal of Machine Learning Research, 5: 361-397.

Park, T., Casella, G. (2008). The Bayesian Lasso. Journal of the American Statistical Association, 103:482, 681-686.

Qi, Z., Cui, Y., Liu, Y., Pang, J-S. (2021). Asymptotic Properties of Stationary Solutions of Coupled Nonconvex Nonsmooth Empirical Risk Minimization. Mathematics of Operations Research. ePub ahead of publication. https://doi.org/10.1287/moor.2021.1198

Rudin, W. (1976). Principles of Mathematical Analysis. 3rd Edition, McGraw-Hill, New York.

Seto, S., Wells, M.T., Zhang, W. (2021). Halo: Learning to prune neural networks with shrinkage. In Proceedings of the 2021 SIAM International Conference on Data Mining (SDM), pages 558–566. SIAM.

Shapiro, A., Xu H. (2007). Uniform laws of large numbers for set-valued mappings and subdifferentials of random functions. Journal of Mathematical Analysis and Applications 352 1390–1399.

Smith, L.N. (2017). Cyclical Learning Rates for Training Neural Networks. arXiv eprint: 1506.01186v6.

Tibshirani, R. (1996). Regression shrinkage and selection via the lasso. J. Royal. Statist. Soc B., 58(1): 267-288.

Tibshirani, R., Saunders, M., Rosset, S., Knight, K. (2005). Sparsity and smoothness via the fused lasso. \textit{J. R. Stat. Soc. Ser. B (Stat. Methodol.)}  67: 91–108.

Trzasko, J., Manduca, A. (2009). Highly undersampled magnetic resonance image reconstruction via homotopic L0-minimization.  IEEE Transactions on Medical Imaging, 28(1): 106-121.

van der Vaart, A.W. (1998). Asymptotic Statistics. Cambridge University Press, New York.

Vettam, S., John, M. (2022). On two recent nonconvex penalties for regularization in machine learning. Results in Applied Mathematics, 14, 100256. doi: 10.1016/j.rinam.2022.100256.

Wang, Y., Zhu, L. (2016). Variable Selection and Parameter Estimation with the Atan Regularization Method. Journal of Probability and Statistics, 2016, 1–12.

Yuan M., Lin Y. (2006). Model selection and estimation in regression with grouped variables. \textit{J. R. Stat. Soc. Ser. B (Stat. Methodol.)} 68: 49–67.

Zhang, C.H. (2010). Nearly unbiased variable selection under minimax concave penalty. Annals of Statistics, 38(2): 894–942.

Zou, H., Hastie T. (2005). Regularization and variable selection via the elastic net. \textit{Journal of the Royal Statistical Society, Series B.} 67: 301-320.

Zou, H. (2006) The Adaptive Lasso and Its Oracle Properties, Journal of the American Statistical Association, 101:476, 1418-1429

\end{document}